\documentclass[11pt]{article}

\usepackage{graphicx}
\usepackage{amsmath,amssymb,amsfonts}
\usepackage{booktabs}
\usepackage{array}
\usepackage{makecell}
\usepackage{algorithm}
\usepackage{algpseudocode}
\usepackage{textcomp}
\usepackage{bm}
\usepackage{float}
\usepackage{hyperref}
\usepackage{natbib}
\usepackage{authblk}

\title{MFVINS: Multiple Fisheye Camera-based Visual Inertial System}

\author[1]{Eunseong Jang}
\author[1]{YuJin Chung}
\author[1]{Sang Jun Lee}
\author[2]{Jihyun Yoon}
\author[1]{HyungGi Jo}

\affil[1]{Division of Electronic Engineering, Jeonbuk National University, Jeonju, South Korea}
\affil[2]{BSTAR Robotics, Inc., Palo Alto, CA, USA}

\date{}

\begin{document}

\maketitle

\begin{abstract}
A simultaneous localization and mapping (SLAM) method using a monocular camera and a low-cost inertial measurement unit (IMU) sensor is an effective way to fulfill a low-cost sensor configuration. Using this sensor configuration, visual-inertial system (VINS) focuses on fusing data from a camera and an IMU sensor to estimate the six degrees-of-freedom (DOF) of the sensor pose. Typically, VINS uses only a single camera as visual input, which lead to problems such as error accumulation due to occlusion, various illumination, and textureless environments. In this paper, we propose a new multiple fisheye camera-based visual-inertial system called MFVINS.

We present an IMU-aided FAST feature tracker for multiple cameras that enables efficient extraction and robust matching of local features. Then, the proposed method filters out outliers caused by fisheye distortion on the normalized image plane. Subsequently, a new reprojection error with physical validity constraints is proposed for bundle adjustment using learning-based depth estimation. The proposed method is applied to various scenarios, and its effectiveness is demonstrated by comparing previous VINS methods. In particular, MFVINS is implemented in real-time process to leverage the advantages of using multiple cameras—robustness against occlusion and textureless regions—while reducing the computational burden.
\end{abstract}


\section{Introduction}
\label{sec1}
Simultaneous Localization and Mapping (SLAM) \cite{campos2021orb, cho2021sp, liu2025transfer} is a core technology that enables robots and autonomous systems to localize themselves while building a map of the surrounding environment. Among various SLAM methods, the visual-inertial system (VINS) \cite{vinsmono} has garnered significant attention for its ability to fuse camera and inertial measurement unit (IMU) data to achieve accurate six degrees-of-freedom (DOF) pose estimation of a camera and maintain trajectory consistency. Most existing VINS solutions adopt a monocular pinhole camera due to its simplicity and cost-effectiveness. However, these systems often suffer from accumulated errors in challenging environments, such as those with occlusions, dynamic lighting conditions, or textureless surfaces. Crucially, using only images with a limited FOV will result in rapid performance degradation when occlusion occurs due to dynamic objects blocking the view.

To overcome these limitations, previous studies have explored the use of wide field-of-view (FoV) cameras and multi-camera setups \cite{mcvio}. While fisheye cameras can capture a broader view of the environment and mitigate occlusion or limited texture issues, they also introduce significant image distortion, complicated feature tracking and depth estimation. Additionally, multi-camera systems inevitably increase computational complexity and often process cameras independently, without fully exploiting the geometric relationships between them. Existing multi-camera systems such as MCVIO \cite{mcvio}, a non-overlapping multi-camera system, handle cameras independently. This characteristic makes it difficult to accurately estimate the scale and depth of extracted features. Even when inter-camera feature matching is achieved, it often occurs in severely distorted areas, which limits accurate depth recovery. 

In this paper, we propose a novel multiple fisheye camera-based visual-inertial system (MFVINS) that addresses the aforementioned challenges by tightly integrating multiple fisheye cameras with a low-cost IMU. Unlike prior systems, MFVINS applies a learning-based depth estimation model to each fisheye image, enabling more precise 3D reconstruction even in visually degraded scenes.

In the frontend of MFVINS, a grid-based FAST algorithm is employed to extract features efficiently, and an IMU-aided feature tracker is implemented to robustly match local features across distorted fisheye images. IMU data can provide the initial locations of feature points, thus tracking efficiency has been enhanced. We also introduce a novel outlier filtering mechanism that identifies and removes mismatched features arising from heavy fisheye distortion by projecting them onto a normalized image plane. Also, deep learning-based depth estimation is exploited to incorporate depth values for subsequent localization. In the backend, loop closure detection and pose graph optimization are performed, significantly reducing long-term drift. 

The proposed system is evaluated on various real-world sequences that include occlusions and low-texture environments. The results demonstrate that MFVINS outperforms existing monocular and multi-camera VINS frameworks in terms of both accuracy and robustness. Moreover, the system is implemented in real-time, showing that the use of multiple fisheye cameras can be practically viable without excessive computational overhead.
The main contributions of this paper are as follows:
\begin{enumerate}
    \item We propose MFVINS, a visual-inertial SLAM system based on multiple fisheye cameras, which improves robustness to occlusion and poor texture while reducing drift through frontend algorithms and backend optimization.
    \item In the frontend, we develop an IMU-aided FAST feature tracking method along with a distortion-aware outlier filtering mechanism to enhance feature robustness under fisheye distortion. Also, an accurate local matching method is proposed by exploiting learning-based depth estimation. 
    \item We validate the system through extensive real-world experiments and ablation studies, demonstrating the effectiveness of each proposed module in improving pose estimation accuracy.
\end{enumerate}

The remainder of this paper is organized as follows. Section 2 reviews related work. Section 3 provides an overview of MFVINS, including notation and coordinate frames. Section 4 details the proposed method and its key components. Section 5 presents the experimental setup, performance comparisons, and ablation analysis. Section 6 concludes and discusses potential future work.

\section{Related Works}

\subsection{Mono/Stereo/RGB-D camera system}{
In VINS, various sensor configurations are employed. VINS-Mono \cite{vinsmono} improves localization performance by tightly coupling a monocular camera and an IMU, along with four-degree-of-freedom pose graph optimization. Similarly, ORB-SLAM3 \cite{campos2021orb} integrates a monocular camera and an IMU and uses Oriented FAST and Rotated BRIEF (ORB) \cite{rublee2011orb} features to enhance real-time performance. However, because a monocular camera has intrinsic limitations, such as scale ambiguity, subsequent research has investigated VIO systems that use stereo cameras. MSCKF-VIO \cite{sun2018robust} tightly couples \cite{forster2016manifold} a stereo camera with an IMU, handling state estimation with a filter-based approach  \cite{mourikis2007multi}.

Beyond relying on parallax from stereo configurations, some VIO methods \cite{ye2026dc, wang2026otps} employ RGB-D cameras (e.g., structured-light or Time-of-Flight sensors) to directly obtain accurate depth data or use learned depth images. VINS-RGBD \cite{shan2019rgbd} extends VINS-Mono by obtaining the depth information of feature points from an RGB-D camera, enabling more precise localization. Dynamic-VINS \cite{liu2022rgb} incorporates object detection algorithms along with RGB-D images, and applies a grid-based efficient FAST feature extraction method to improve computational efficiency. Meanwhile, CodeVIO \cite{zuo2021codevio} incorporates depth images obtained from a learning-based network into conventional pinhole images for VIO.
}

\begin{table*}[ht]
\centering
\footnotesize
\resizebox{1\textwidth}{!}{%
\begin{tabular}{lccccc}
    \hline
    & VINS-Mono \cite{vinsmono} & VINS-RGBD \cite{shan2019rgbd} & Dynamic-VINS \cite{liu2022rgb} & MCVIO \cite{mcvio} & MFVINS(Ours) \\
    \hline
    Multicam (\(\geq\) 2 cameras) & x & x & x & o & o \\
    Feature extraction & Shi-Tomasi & Shi-Tomasi & FAST & \makecell{Shi-Tomasi\\VPI-harris} & \makecell{Shi-Tomasi\\VPI-harris\\FAST} \\
    IMU aided feature tracker & x & x & o & x & o \\
    Depth Image & x & o & o & x & o \\
    PGO (loop closure) & o & o & o & x & o \\
    Relocalize & o & o & o & x & o \\
    Fisheye & o & o & o & o & o \\
    \hline
\end{tabular}%
}
\caption{Comparison of VINS-based systems.}
\label{tab:comparison}
\end{table*}

\subsection{Multiple camera system}{
Using multiple cameras can improve both the accuracy and robustness of VINS. BAMF-SLAM \cite{zhang2023bamf} utilizes a combination of monocular and stereo fisheye cameras and employs Recurrent Field Transforms (RFT) and Bundle Adjustment (BA) to enhance the accuracy of a VI-SLAM system. ROVINS \cite{seok2020rovins} integrates an omnidirectional fisheye camera with an IMU to increase the robustness of pose estimation. OmniNxt \cite{liu2024omninxt} also uses an omnidirectional fisheye camera, generating depth images via a CNN that employs cylindrical warping and stereo matching. These depth images are then used for more accurate localization. However, since the experiments were limited to relatively small environments, the scalability to large-scale scenarios remains unverified. Furthermore, many such systems like MAVIS \cite{wang2024mavis} with a front stereo sensor and side monocular cameras require specific sensor configurations.

Table \ref{tab:comparison} summarizes the key functional comparisons between the proposed MFVINS and representative visual-inertial systems (\cite{mcvio,vinsmono,shan2019rgbd,liu2022rgb}). These existing methods employ diverse sensor configurations, including a single camera, RGB-D sensors, and multi-camera setups, and differ in aspects such as feature extraction algorithms (e.g., Shi-Tomasi\cite{harris1988combined}, FAST\cite{viswanathan2009features}), the use of IMU-based tracking, and the incorporation of depth information. While MCVIO \cite{mcvio} supports a multi-camera arrangement, it processes the features from each camera independently, thus failing to fully leverage the geomteric information between cameras and limiting feature depth estimation. Furthermore, neither the publicly available code nor the experiments presented in the paper include loop closure or pose graph optimization to correct cumulative errors, making the system prone to drift over extended tracking periods. 
}

\section{Overview}

\begin{figure*}
\centering
\includegraphics[width=1\textwidth]{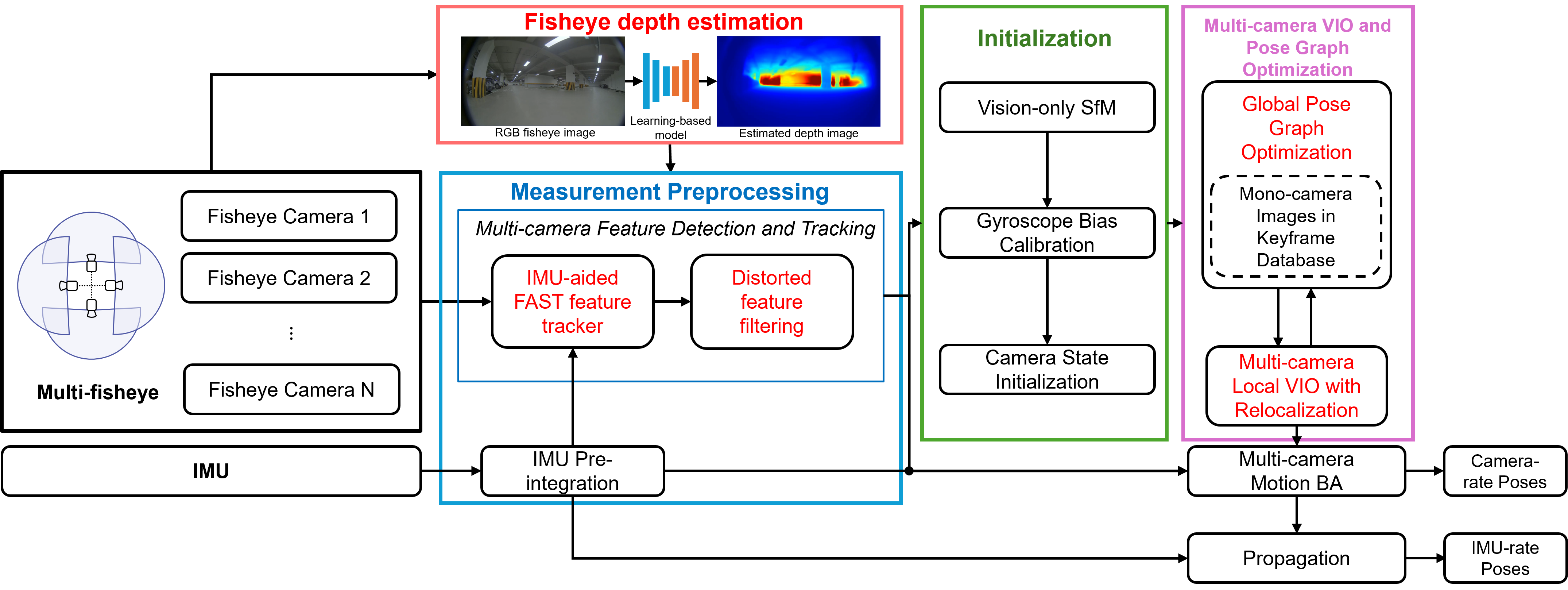}
\caption{Framework of the proposed Multiple Fisheye Visual Inertial Systems (MFVINS). The upgraded modules are highlighted in red.}
\label{fig:Framework}
\end{figure*}

\subsection{Proposed Frameworks}
The proposed localization framework builds upon the foundational structures of VINS-Mono \cite{vinsmono} and MCVIO \cite{mcvio}. As illustrated in the block diagram of Fig. \ref{fig:Framework}, the system is composed of four primary components: 1) Measurement Preprocessing, 2) Fisheye depth estimation, 3) Initialization, 4) Multi-Cam VIO and Pose Graph Optimization (PGO) \cite{carlone2015initialization}. The core contributions of our method are highlighted in red within the Fig. \ref{fig:Framework}.

\subsubsection{Measurement Preprocessing}
In this stage, images collected from multiple fisheye cameras are processed to extract features using the FAST \cite{viswanathan2009features} algorithm. Simultaneously, angular velocity measurements ${{\bm{\omega }}^b}$ obtained from the IMU are used to enhance tracking. Due to potential discrepancies in observation timing between the camera and the IMU, a temporal offset $\Delta {t_{cam - imu}}$ may arise, which is estimated and compensated through calibration for accurate sensor synchronization. By computing a predicted rotation matrix ${{\bf{R}}_{pred}}$ from the calibrated data, the efficiency of feature tracking is significantly improved. Furthermore, feature outliers caused by distortion of fisheye lens are filtered, resulting in a refined set of features ${\mathbf{p}}_i$ that contributes to the reliability and accuracy of downstream processes.

\subsubsection{Fisheye Depth Estimation}
In this module, fisheye images are passed through a deep learning-based depth estimation network to obtain either pixel-wise depth maps $\mathbf{D}$  or inverse depth values $\bm{\lambda}$ . The resulting depth information is then associated with each extracted feature from the preprocessing stage, providing essential geometric constraints for accurate localization. 

\subsubsection{Initialization}
For system initialization, a structure-from-motion (SfM) \cite{schonberger2016structure} technique is applied using primarily the front-facing camera. This step establishes the initial position and orientation of the robot (or camera) and estimates the necessary parameters for initializing visual-inertial odometry (VIO). 

\subsubsection{Multi-Cam VIO and Pose Graph Optimization (PGO)}
During this phase, visual-inertial odometry is performed using measurements from the multi-fisheye camera setup. The state of the system is updated using a sliding-window optimization strategy. To mitigate drift errors accumulated over time, loop closure detection and pose graph optimization are applied using the front camera, which consistently observes the forward motion direction. This approach enhances both localization accuracy and long-term stability.

\subsection{Notations and frame definitions}
{
To maintain consistency with VINS-Mono \cite{vinsmono}, the notations and coordinate frames used in this study follow the same conventions. We define the world frame $(\cdot)^w$, camera frame $(\cdot)^c$, IMU body frame $(\cdot)^b$, and an additional LiDAR frame $(\cdot)^l$ . The transformation between these coordinate frames is represented by a 4×4 matrix in the SE(3) group, defined as
\begin{equation}
\label{Transform}
{\bf{T}} = \left[ {\begin{array}{*{20}{c}}
{\bf{R}}&{\bf{t}}\\
{{{\bf{0}}^T}}&1
\end{array}} \right]\; \in {\rm{SE(3)}}
\end{equation}
where ${\bf{t}} \in {\mathbb{R}^3}$ is the translation vector, and ${\bf{R}} \in {\rm{SO(3)}}$  is the rotation matrix, which can also be represented as a rotation vector or quaternion as 

\begin{equation}
\label{so3}
{\bf{r}} = {\left( {\left[ {\begin{array}{*{20}{c}}
{{r_1}}\\
{{r_2}}\\
{{r_3}}
\end{array}} \right]} \right)_ \times } = \left[ {\begin{array}{*{20}{c}}
0&{ - {r_3}}&{{r_2}}\\
{{r_3}}&0&{ - {r_1}}\\
{ - {r_2}}&{{r_1}}&0
\end{array}} \right] \in \mathfrak{so}(3)
\end{equation}

Based on these coordinate settings, we define two projection functions, ${\pi _l}\left(  \cdot  \right):{\mathbb{R}^3} \to {\mathbb{R}^2}$  and ${\pi _0}\left(  \cdot  \right):{\mathbb{R}^3} \to {\mathbb{R}^2}$, to map a 3D point ${{\bf{P}}^w} \in {\mathbb{R}^3}$  in the world frame to the 2D image plane. ${\pi _l}\left(  \cdot  \right)$  corresponds to projection through a fisheye camera model \cite{kannala2006generic}, while ${\pi _0}\left(  \cdot  \right)$  uses a unit sphere projection.
Let ${{\bf{p}}_n} = {\left[ {u',v'} \right]^T}$ be the projected 2D point on the normalized image plane corresponding to a 3D point ${\bf{P}}^w$. The radial distance from the camera center to $\left( {u',v'} \right)$ is denoted as $r' = \sqrt {{{u'}^2} + {{v'}^2}}$. The incident angle in the normalized coordinate system is defined as $\theta$, and the distorted incident angle that accounts for radial distortion is denoted as $\theta_d$. The distortion is modeled using radial distortion coefficients ${k_1},{k_2}, \cdots$. Under the fisheye camera model, the projection function ${\pi _l}\left(  \cdot  \right)$ that maps a 3D point onto the image plane is given by
\begin{equation}
\label{theta-d}
{\theta _d} = \theta  + {k_1}{\theta ^3} + {k_2}{\theta ^5} +  \cdots
\end{equation}
\begin{equation}
\label{pi-l}
{\pi _l}\left( {\bf{P}} \right) = \left[ {\begin{array}{*{20}{c}}
{{f_x} \cdot {\theta _d}\left( {\frac{{u'}}{{r'}}} \right) + {c_x}}\\
{{f_y} \cdot {\theta _d}\left( {\frac{{v'}}{{r'}}} \right) + {c_y}}
\end{array}} \right]
\end{equation}
where $f_x, f_y$  are the focal lengths, and $c_x, c_y$  is the principal point of the camera.

In the case of directly projecting a 3D point ${\bf{P}} = {\left[ {{P_x}\,\,{P_y}\,\,{P_z}} \right]^T}$ located on the unit sphere, the incident angle $\theta  = {\cos ^{ - 1}}\left( {\frac{{{P_z}}}{{\left\| {\bf{P}} \right\|}}} \right)$ and azimuth angle $\phi  = {\tan ^{ - 1}}\left( {\frac{{{P_y}}}{{{P_x}}}} \right)$ are first computed. Similar to the previous model, a distorted incident angle $\theta_d$ is applied to account for radial distortion using the same distortion model. The projection function ${\pi_0}\left(  \cdot  \right)$ , which maps a unit-sphere point to the image plane, is defined as follows
\begin{equation}
\label{pi-0}
{\pi _0}\left( {\bf{P}} \right) = \left[ {\begin{array}{*{20}{c}}
{{f_x} \cdot {\theta _d} \cdot \cos \phi  + {c_x}}\\
{{f_y} \cdot {\theta _d} \cdot \sin \phi  + {c_y}}
\end{array}} \right].
\end{equation}

The acceleration  ${\bf{\hat a}}_t^b$ and angular velocity ${\bf{\hat \omega }}_t^b$ measured by the IMU at time t are expressed as
\begin{equation}
\label{measure}
\begin{split}
\hat{\mathbf{a}}_t^b &= \mathbf{a}_t^b + \mathbf{b}_{a_t} + \mathbf{R}_w^t \mathbf{g}^w + \mathbf{n}_a \\
\hat{\boldsymbol{\omega}}_t^b &= \boldsymbol{\omega}_t + \mathbf{b}_{\omega_t} + \mathbf{n}_\omega
\end{split}
\end{equation}
where ${{\mathbf{b}}_{{a_t}}}$ and ${{\mathbf{b}}_{{\omega _t}}}$ are the accelerometer and gyroscope biases, respectively, and ${{\mathbf{n}}_a},{{\mathbf{n}}_\omega }$ denote Gaussian white noise. To handle these biases and noise components during state estimation, T. Qin, \textit{et al.}  \cite{vinsmono} proposed an IMU preintegration method. In this paper, we adopt the notational conventions from Y. He, \textit{et al.} \cite{mcvio}.
}

\begin{figure}
\centering
\includegraphics[width=1\textwidth]{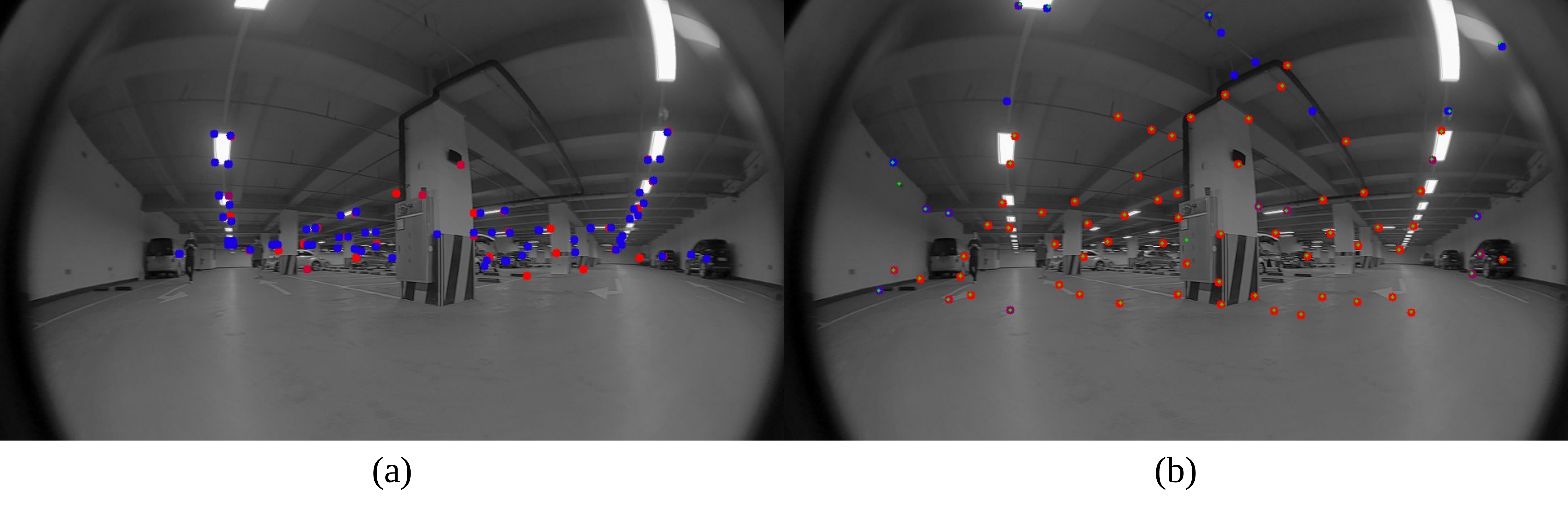}
\caption{An example of feature detection and tracking (Blue: detected, Red: tracked, Green: predicted features) (a) MCVIO using VPI-Harris, (b) proposed method.}
\label{fig:_feature_extrator_compare}
\end{figure}

\section{Multiple Fisheye VINS(MFVINS)}
\subsection{IMU-aided FAST feature tracker}
\subsubsection{Visual Feature Extraction}
Fisheye cameras offer a significantly wider field of view (FoV) compared to conventional pinhole cameras, resulting in the detection of a larger number of visual features. In multi-fisheye camera systems, this number increases further, posing computational challenges. Therefore, lightweight and efficient feature detection becomes crucial for real-time performance.

Our proposed method is modified version of an IMU-aided FAST feature tracking method proposed in \cite{liu2022rgb}, tailoring it for multi-fisheye camera configurations. In our implementation, the image is divided into an N×M grids, and FAST features are uniformly extracted within each grid cell to ensure even spatial coverage, as illustrated in Fig. \ref{fig:_feature_extrator_compare}(b). Parallel threading is also used to accelerate computation. This process is independently applied to each fisheye camera, enabling scalable and efficient multi-camera feature extraction suitable for real-time SLAM operations.

\subsubsection{IMU-aided Tracking}

\begin{figure}
\centering
\includegraphics[width=1\textwidth]{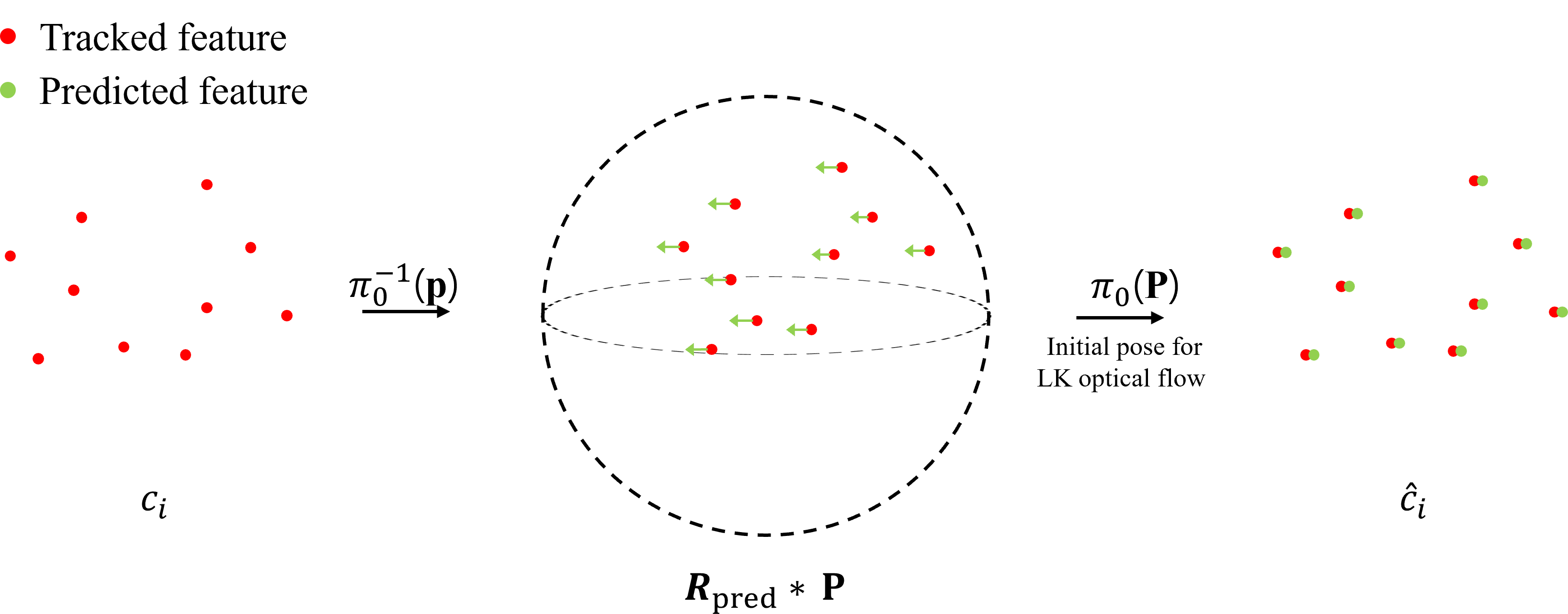}
\caption{Predicting feature point locations using a rotation matrix}
\label{fig:pred_feature}
\end{figure}

Following the feature extraction step, feature tracking is typically performed using the Lucas-Kanade (LK) optical flow algorithm. However, fisheye images inherently exhibit strong radial distortion and often contain a number of features, which can degrade the accuracy and efficiency of conventional feature matching techniques.

Firstly, a key challenge in integrating IMU and camera data is the presence of temporal misalignment due to differing sensor acquisition time. To mitigate this, our method estimates and compensates for the temporal offset $\Delta {t_{cam - imu}}$  between the IMU and each camera. Unlike \cite{liu2022rgb}, which only considers the offset in the current frame, our method corrects the offset for both the current frame $b_{k+1}$ and previous frames $b_{k}$ to improve the precision of the synchronization.

Using the compensated angular velocity readings from the IMU, the rotation between two consecutive frames is estimated. The 2D features ${{\bf{p}}_{{c_i}}} = \left[ {{u_j},{v_j}} \right]_{j = 1:{N_p}}^T$ are back-projected onto the unit sphere ${{\bf{P}}_{{c_i}}}$ using the fisheye model, recovering their 3D directions. In the fisheye camera model, from azimuth $\phi  = {\tan ^{ - 1}}\left( {\frac{{v'}}{{u'}}} \right)$ and distance $r' = \sqrt {{{u'}^2} + {{v'}^2}}$ between the camera origin and $\left( {u',v'} \right)$, thus $\left( {u',v'} \right) = \left( {\frac{{{\theta _d}}}{r}u,\frac{{{\theta _d}}}{r}v} \right)$. Therefore, $r'$ exactly corresponds to the distorted incidence angle theta, and (\ref{theta-d}) can be re-written as follows
\begin{equation}
\label{theta}
\theta (1 + {k_1}{\theta ^2} + {k_2}{\theta ^4} + {k_3}{\theta ^6} + {k_4}{\theta ^8}) - r' = 0.
\end{equation}
        
Then, the undistorted incident angle $\theta$ is computed by solving a monic polynomial (\ref{theta}) whose roots correspond to the true angle of incidence. To solve (\ref{theta}), the companion matrix is defined as

\begin{equation}
C = \left[ {\begin{array}{*{20}{l}}
0&0&0& \cdots &0&0&{r'}\\
1&0&0& \cdots &0&0&{ - 1}\\
0&1&0& \cdots &0&0&0\\
0&0&1& \cdots &0&0&{ - {k_1}}\\
 \vdots & \vdots & \vdots & \ddots & \vdots & \vdots & \vdots \\
0&0&0& \cdots &1&0&0\\
0&0&0& \cdots &0&1&{ - {k_4}}
\end{array}} \right] \in {\mathbb{R}^{10 \times 10}}.    
\end{equation}

The roots of (\ref{theta}) are obtained by computing the eigenvalues of $C$. We apply \textit{Schur decomposition} \cite{wigderson2019mathematics} to the $C$ as
\begin{equation}
C=TUT^{-1},    
\end{equation}
where $T$ is unitary matrix and $U$ is upper triangular matrix. Then, the diagonal values of $U$ are eigenvalues, and the positive values are selected from the real eigenvalues with sufficiently small imaginary parts, and the smallest value among them is determined as the final restored $\theta$. Using the recovered incident angle $\theta$ and the computed azimuth angle $\phi$, the corresponding 3D point on the unit sphere can be reconstructed as
	 	
\begin{equation}
\label{(pi-o-inv)}
{{\bf{P}}_{{c_i}}^{c}} = \pi _0^{ - 1}\left( {{{\bf{p}}_{{c_i}}}} \right) = \left[ {\begin{array}{*{20}{c}}
{\sin \theta \cos \phi }\\
{\sin \theta \sin \phi }\\
{\cos \theta }
\end{array}} \right].
\end{equation}

The unit vector ${{\bf{P}}_{{c_i}}^{c}}$ represents the direction of the visual feature relative to the $i$-th camera center, compensating for radial distortion introduced by the fisheye lens. 

Next, to estimate the camera motion between two consecutive frames $b_{k}$ and $b_{k+1}$, we integrate the angular velocity sequence $\bm{\omega}^b_t$ measured by the IMU over the time interval $\left[ {{t_k},{t_{k + 1}}} \right]$. The incremental rotation 
${\bm{\delta }}$, represented as an angle-axis vector, is computed as 

\begin{equation}
{\bm{\delta }} = \int_{{t_k}}^{{t_{k + 1}}} {\left( {{\bm{\omega }}_t^b - {{\bf{b}}_g}} \right)} dt.
\end{equation}

In this work, trapezoidal integration is used for improved numerical accuracy, computing the rotation as
\begin{equation}
{\bm{\delta }} \approx \sum\limits_{i = 1}^{N - 1} {\frac{{\Delta {t_i}}}{2}} \left[ {{{\bf{\omega }}_i} - {{\bf{\omega }}_{i + 1}}} \right] - {{\bf{b}}_g}.
\end{equation}
The resulting rotation vector is then converted into a rotation matrix ${{\bf{R}}_{pred}}$
\begin{equation}
\label{R-pred}
{{\bf{R}}_{pred}} = \prod\limits_l {\exp \left( {ext_{{\bf{R}}_{{c_i}}^b}^T{{\left[ { {{\bm{\delta }}_l}} \right]}_ \times }} \right)}.
\end{equation}
where $ext_{{\bf{R}}_{{c_i}}^b}^T$ represents transformation between IMU and $i$-th camera ${c_i},\,i \in \left\{ {front,rear,left,right} \right\}$ and $\mathbf{b}_g$ is bias of IMU gyroscope. (\ref{R-pred}) explains that the accumulated rotation vector obtained by integrating the IMU angular velocity over time is converted into a rotation matrix through the exponential map exp$(\cdot)$ \cite{ma2004invitation}. These rotated feature directions are reprojected onto the image plane using the fisheye projection function.

In summary, as illustrated in Fig. \ref{fig:pred_feature}, each feature point is back-projected to a 3D unit sphere coordinate ${{\bf{P}}_{{c_i}}^{c}} = \pi _0^{ - 1}({{\bf{p}}_{{c_i}}})$  using (\ref{(pi-o-inv)}), assuming a unit distance. After that, the estimated rotation amount ${{\bf{R}}_{pred}}$ is reflected by applying ${{\bf{\hat P}}_{{c_i}}^{c}} = {{\bf{R}}_{pred}}{{\bf{P}}_{{c_i}}^{c}}$, and then converted to 2D image coordinates ${{\bf{\hat p}}_{{c_{i + 1}}}}$ through the projection function ${\pi _0}({{\bf{\hat P}}_{{c_i}}^{c}})$. The calculated position is used as the initial position of the LK optical flow \cite{lucas1981iterative}, contributing to improving both the accuracy and convergence rate of feature matching, especially under large rotations and fisheye distortion.

\subsubsection{Outlier Filtering for Fisheye-Induced Distortion}

Visual features are extracted and tracked using the Kanade-Lucas-Tomasi (KLT) sparse optical flow algorithm \cite{klt}. To remove mismatched correspondences of features, RANSAC \cite{ransac} and the fundamental matrix \cite{fundamental} are used during the preprocessing stage, along with distortion compensation techniques. While these methods are effective under mild distortion, they become less reliable when applied to fisheye images with non-square aspect ratios (e.g., 16:9), as opposed to 1:1 image commonly used in datasets such as TUM VI \cite{schubert2018tum}. 

\begin{figure}
\centering
\includegraphics[width=0.8\textwidth]{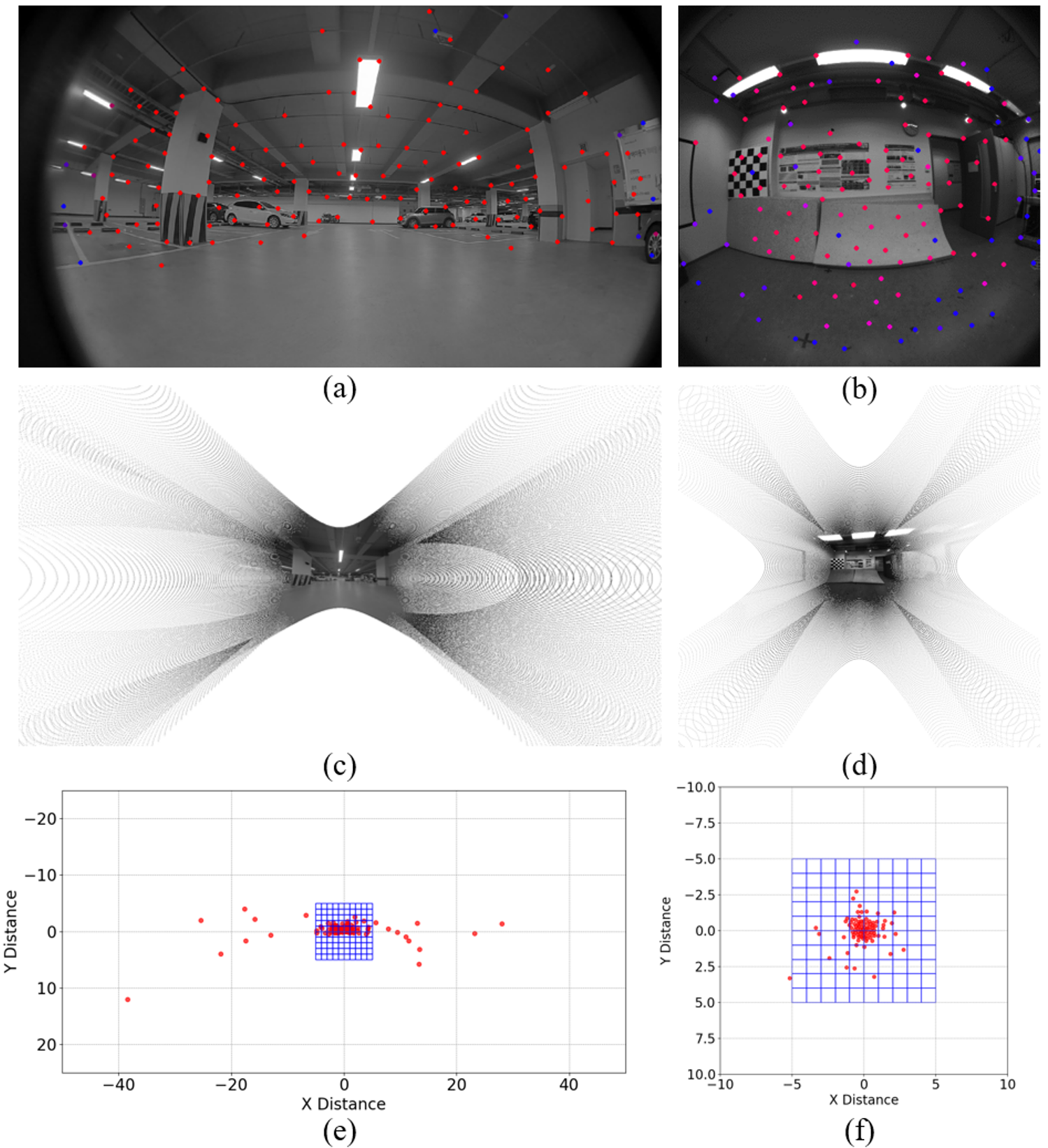}
\caption{Comparison of feature distributions on the normalized image plane between TUM VI and our 16:9 fisheye dataset.
(a), (b): Tracked feature points on the raw fisheye images.
(c), (d): Distortion-corrected fisheye images.
(e), (f): Distribution of tracked features projected onto the normalized plane.
}
\label{fig:compare_undist}
\end{figure}

Fig. \ref{fig:compare_undist} compares feature tracking results from our 16:9 fisheye camera setup and the TUM VI dataset. Fig. \ref{fig:compare_undist}(a) and \ref{fig:compare_undist}(b) show tracked feature points, Fig. \ref{fig:compare_undist}(c) and 5\ref{fig:compare_undist}(d) present distortion-corrected images, and Fig. \ref{fig:compare_undist}(e) and \ref{fig:compare_undist}(f) display the distribution of tracked features on the normalized image plane. In the TUM VI case (1:1 aspect ratio), features are concentrated near the center and symmetrically distributed, with all tracked points falling within the 10×10 grid boundary, as shown in Fig. \ref{fig:compare_undist}(f). However, for our 16:9 images, fisheye distortion causes a horizontal spreading of feature points, resulting in outliers that extend beyond the normalized boundary, as clearly visible in Fig. \ref{fig:compare_undist}(c) and \ref{fig:compare_undist}(e). These observations highlight the need for a more robust outlier rejection method that can effectively handle distortion-specific artifacts, particularly when using wide FoV fisheye cameras with non-standard aspect ratios.

To address this issue, a distortion-aware outlier filtering method is proposed that operates on the normalized image plane. Let ${{\bf{p}}_{n,j}} \in {\mathbb{R}^2}$ denote the normalized 2D coordinates of the $j$-th tracked feature in a given frame. For each tracked feature, we compute its Euclidean norm ${\left\| {{{\bf{p}}_{n.j}}} \right\|_2}$ to assess its deviation from the center of the normalized plane.

A threshold $\tau$ is defined to eliminate feature points that exhibit excessive radial displacement due to fisheye distortion. In our implementation, we empirically set $\tau=6$. The filtering condition is defined as
\[
{{\bf{p}}_{f}} = \left\{ {{{\bf{p}}_{n.j}}\left| {{{\left\| {{{\bf{p}}_{n.j}}} \right\|}_2} > \tau ,\;\;j = 1, \ldots ,{N_p}} \right.} \right\}.
\]

\begin{figure}
\centering
\includegraphics[width=0.7\textwidth]{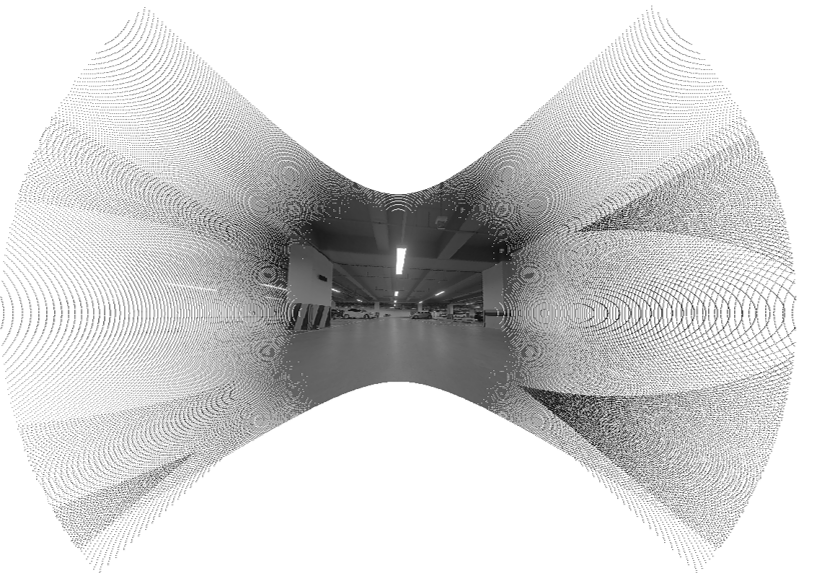}
\caption{Distortion-corrected image after applying the proposed filtering method}
\label{fig:undist}
\end{figure}

Applying this filter effectively removes features located in the diverging peripheral regions of the distorted image, particularly in wide aspect ratio fisheye frames. As shown in Fig. \ref{fig:undist}, this method eliminates horizontally diverging outliers while preserving features within the valid central region of the normalized plane.

\subsubsection{Learning-based Depth Estimation and Multi-Camera Visual Inertial Odometry}

While RGB-D cameras provide direct depth measurements, they often suffer from limitations in large-scale environments \cite{darwish2019robust}. Most RGB-D sensors are designed for short-range pinhole configurations, resulting in decreased depth accuracy at longer distances and poor compatibility with wide-FoV lenses. Similarly, traditional stereo matching methods encounter difficulties under fisheye distortion, leading to unreliable triangulation results.

\begin{figure}
\centering
\includegraphics[width=0.8\textwidth]{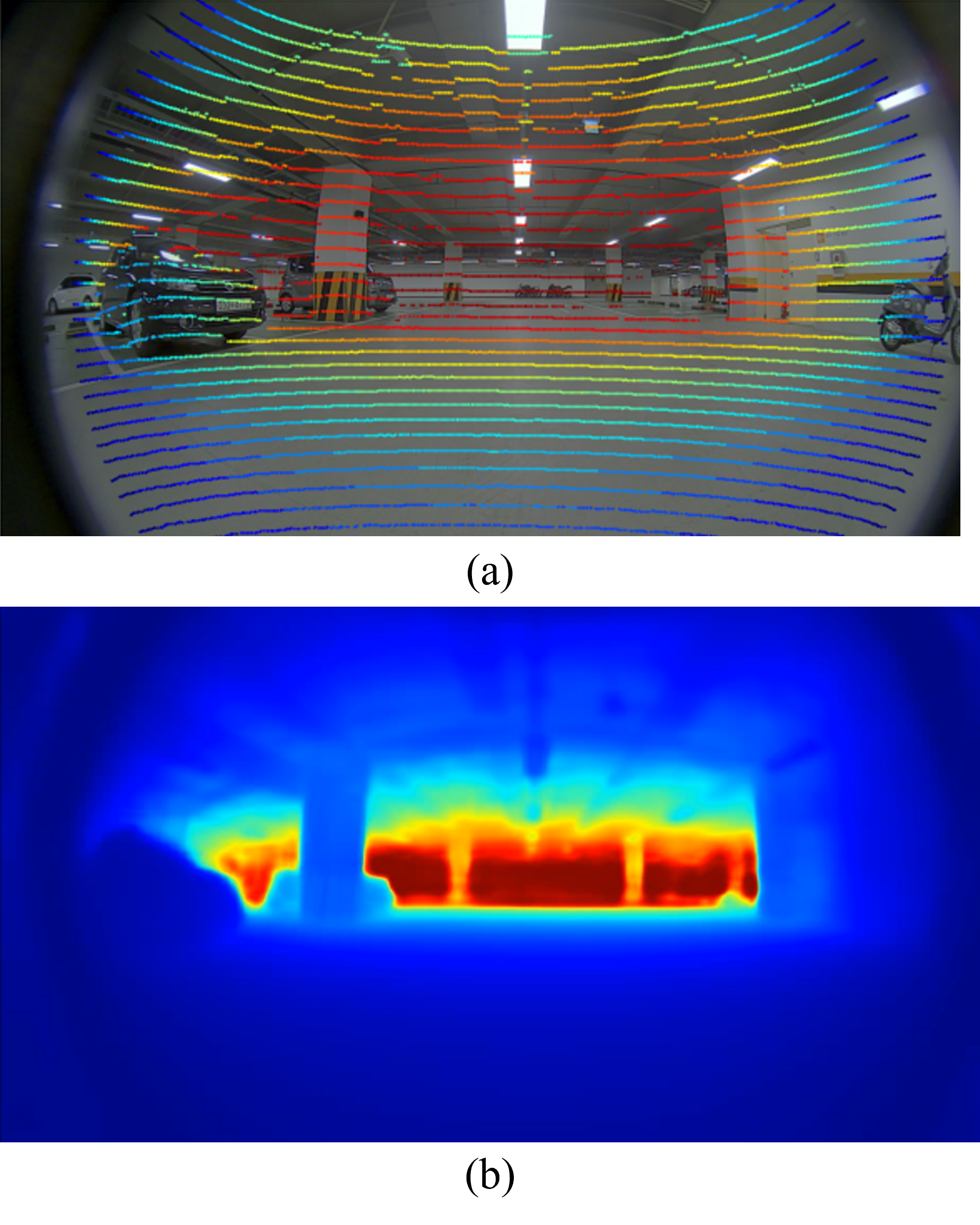}
\caption{(a) Fisheye image with projected point cloud, (b) Deep learning-based depth image}
\label{fig:deep_depth}
\end{figure}

To address these challenges, a learning-based depth estimation module tailored to our multi-fisheye camera setup is implemented. Specifically, LiDAR-generated point clouds are projected onto each fisheye image using the camera model $\pi_l$ as shown in Fig. \ref{fig:deep_depth}(a). Then, training data can be generated that reflects lens distortion. Using projected point clouds, the monocular depth estimation model \cite{lee2019big} is fine-tuned, initially trained on the KITTI dataset \cite{geiger2012we}. This model outputs a dense inverse depth map ${{\bf{D}}_i}$, as illustrated in Fig. \ref{fig:deep_depth}(b).

The predicted inverse depth ${\hat{\lambda}_{i,j}}$ of the features ${{\bf{p}}_f} = \left[ {{u_j},{v_j}} \right]_{j = 1:{N_p}}^T$ is selectively incorporated into the VIO pipeline. The learning-based depth map is adopted as an initialization if it satisfies physical validity constraints. Otherwise, conventional triangulation is used as
\begin{equation}
{\hat \lambda _{i,j}} = \left\{ {\begin{array}{*{20}{l}}
{{{\bf{D}}_i}\left( {{u_{_j}},{v_{{_j}}}} \right),}&{{\rm{if }}{\mkern 1mu} \;0 < {{\bf{D}}_i}\left( {{u_{_j}}},{v_{{_j}}} \right) < \delta }\\
{{\lambda _{i,j}},}&{{\rm{otherwise}.}}
\end{array}} \right.    
\end{equation}

Then, the depth of feature is refined through optimization using the Singular Value Decomposition (SVD) approach to minimize reprojection errors, 

\begin{equation}
\begin{array}{l}
{{\hat \lambda }_{i,j}}{{\bf{p}}_{f,j + 1}} = {\bf{R}}_j^{j + 1}\left( {{{\hat \lambda }_{i,j}}{{\bf{p}}_{f,j}}} \right) + {\bf{t}} + {{\bf{e}}_j}\\
\sum {{\bf{e}}_j^2}  = \sum\limits_{j = 1}^{{m_i}} {{{\left\| {{{\hat \lambda }_{i,j}}{{\bf{p}}_{f,j + 1}} - {\bf{R}}_j^{j + 1}\left( {{{\hat \lambda }_{i,j}}{{\bf{p}}_{f,j}}} \right) - {\bf{t}}} \right\|}^2}} .
\end{array}
\end{equation}
This scheme reduces the risk of optimization contamination in sections where deep learning results produce excessively large errors, while actively utilizing depth information in areas with high prediction stability to compensate for depth estimation uncertainty that tends to occur in wide-angle (fish-eye) camera environments. As a result, this system presents an integrated approach that improves accuracy and stability by incorporating deep learning.

\subsection{Pose Graph Optimization for Drift Correction}
To mitigate drift accumulated during visual-inertial odometry, we adopt a back-end pipeline similar to that of VINS-Mono \cite{vinsmono}. Specifically, loop closure detection is performed using DBoW2 \cite{galvez2012bags} on images captured from the front-facing camera. When a loop is detected, relocalization is triggered to correct accumulated pose errors.
Following loop detection, pose graph optimization (PGO) is applied to enforce global consistency across the trajectory. This optimization step reduces long-term drift by aligning revisited locations while preserving local odometry accuracy. In our system, only the front camera is used for loop closure detection and pose graph optimization. This method not only increases the probability of loop detection by stably observing the scene in the driving direction, but also reduces computational complexity, thereby securing real-time performance.

\section{Experiments}

\begin{figure}
\centering
\includegraphics[width=1\textwidth]{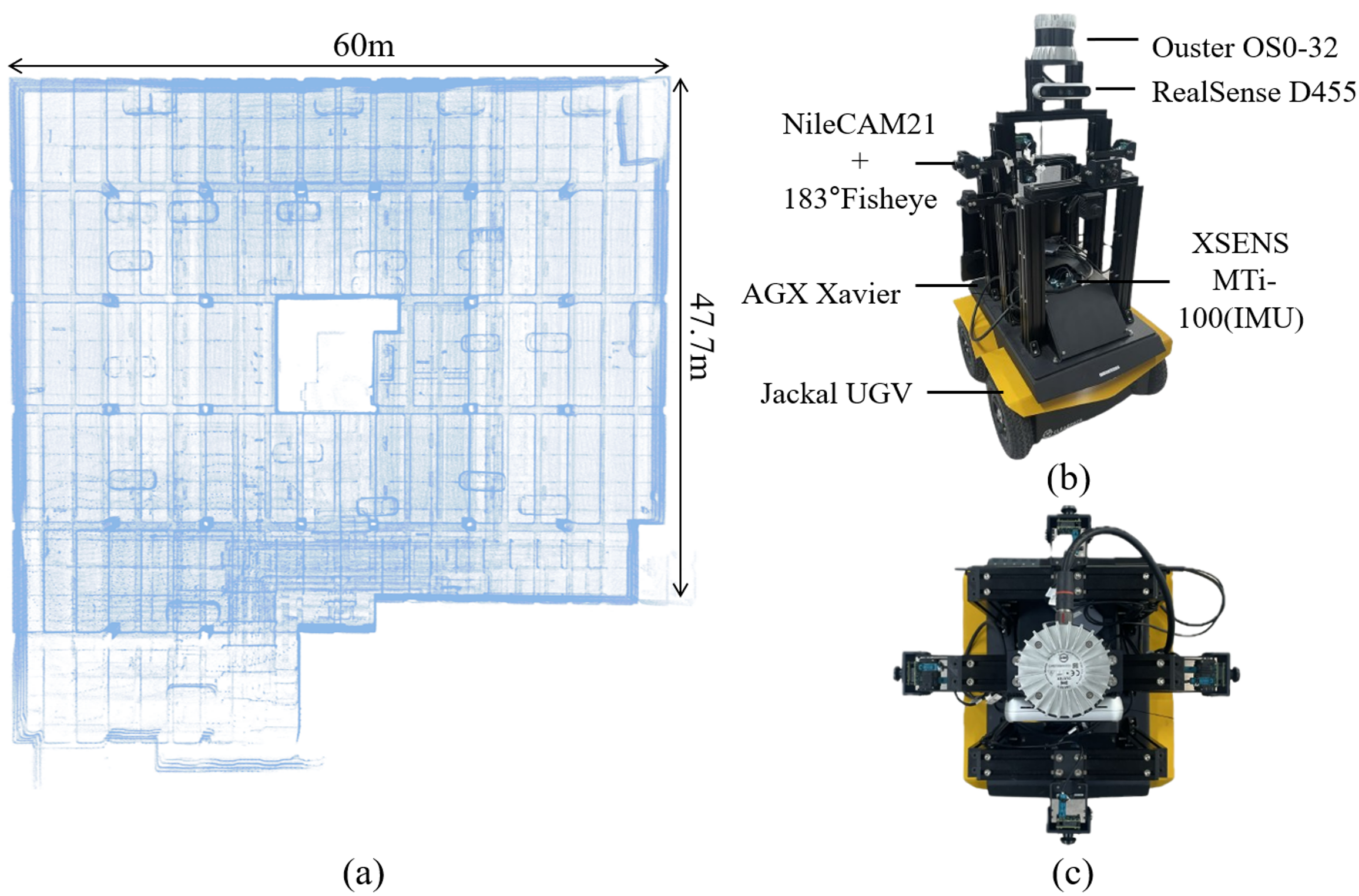}
\caption{(a) 3D point cloud map of underground parking lot where experiments conducted; (b), (c) Side and top views of the mobile robot platform equipped with multiple sensors, respectively.}
\label{fig:env}
\end{figure}

\begin{figure}
\centering
\includegraphics[width=1\textwidth]{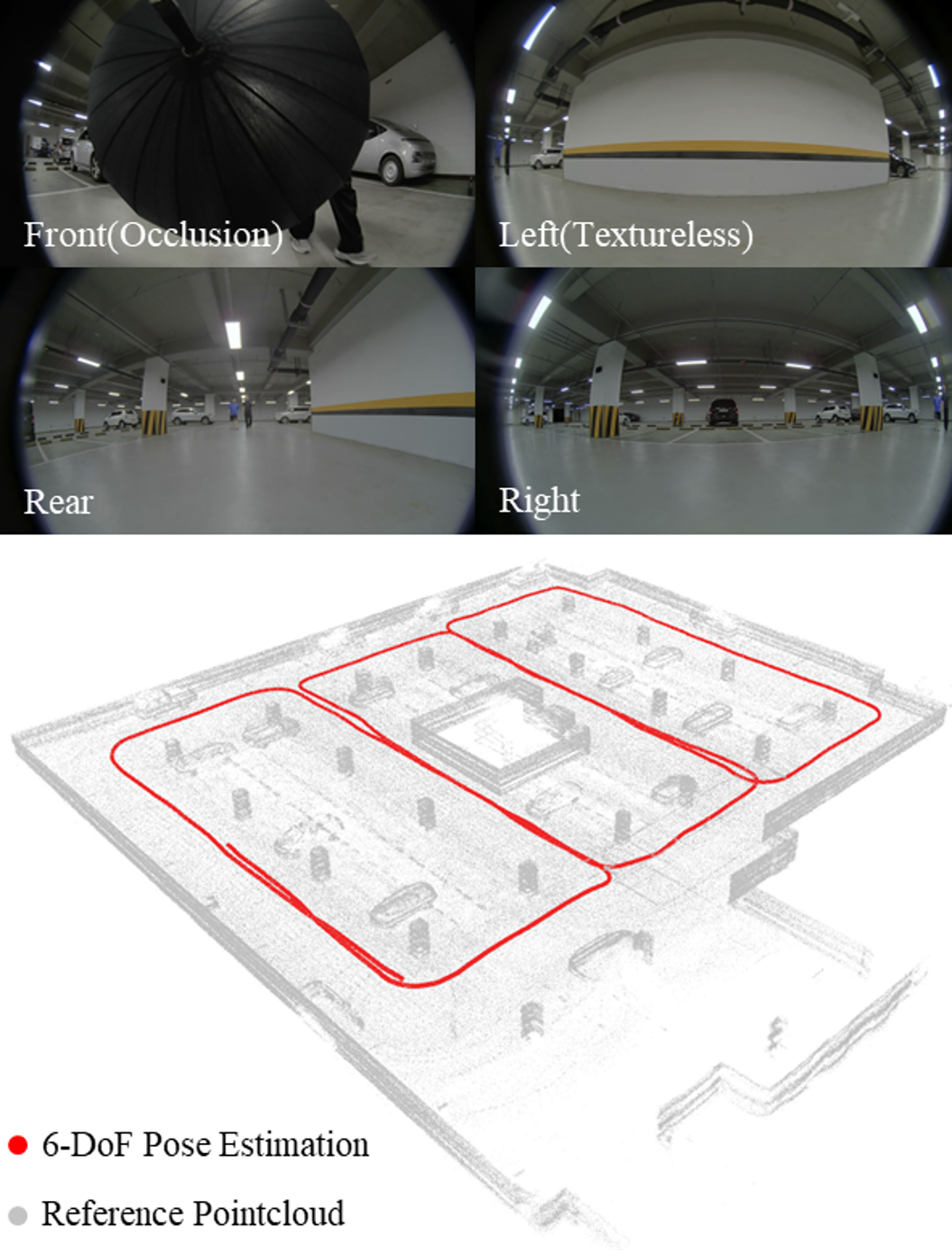}
\caption{A sample data in an underground parking lot. The top panel shows multi-camera sensor data captured from the robot platform, highlighting challenging conditions such as occlusion (front) and textureless surfaces (left). The bottom panel illustrates the generated 3D point cloud map and the estimated 6-degree-of-freedom (6-DoF) pose trajectory (red) aligned with the reference point cloud (gray).}
\label{fig:env2}
\end{figure}

In this paper, to evaluate the performance of localization in a large-scale indoor environment, we conducted experiments in the underground parking lot. Three different scenarios were considered: \textit{DRIVING}, \textit{OCCLUSION}, and \textit{ROTATIONAL}. We used a two-wheeled UGV platform, shown in Figs. \ref{fig:env}(b) and (c). Four fisheye cameras with a 183° FoV are installed to gather omnidirectional view, along with an IMU (Xsens MTi-100). A 3D LiDAR (Ouster OS0-32) is attached to acquire ground-truth for evaluation. To compare performance with a pinhole camera setup, we also installed a RGB-D sensor (Intel RealSense D455). All sensors were operated on an embedded board (NVIDIA Jetson AGX Xavier). As illustrated in Fig. \ref{fig:env2}, the test environment presents significant operational challenges, characterized by frequent occlusions in the front-facing camera and textureless surfaces (e.g., plain walls in the left view) that typically cause feature tracking failures in standard VINS. Please watch the video clip provided as supplemental material.

\subsection{Evaluation Metrics}
To analyze the performance of the proposed method, we use the Root Mean Square Error (RMSE), which is widely used in evaluations of localization algorithms. RMSE is used as an Absolute Trajectory Error (ATE). As reference, we use the 6-DoF pose of a LiDAR-based localization algorithm \cite{bai2022faster} as ground-truth, aligning the trajectory of the visual-inertial system with this reference to evaluate accuracy.

Ablation studies were conducted to evaluate the performance of each component of the proposed method. In this process, both the RMSE and several feature-related metrics were assessed. The evaluated metrics include \textit{feature detection time}, \textit{the number of untracked features}, and \textit{the average number of frames per tracked feature}. Feature detection time refers to the average time required to detect features between consecutive frames. The number of untracked features indicates the number of features that were detected but not successfully tracked during a 100-second period. Lastly, the average number of frames per tracked feature represents the number of frames during which each successfully tracked feature was maintained on average.

\begin{table*}[t]
\centering
\small
\resizebox{1\textwidth}{!}{%
\begin{tabular}{llccccccccc}
\toprule
Sequence & &
\multicolumn{6}{c}{\textit{DRIVING}} & 
\multicolumn{3}{c}{\textit{OCCLUSION}} \\
\cmidrule(lr){3-8}\cmidrule(lr){9-11} 
 &  & Case 1 & Case 2 & Case 3 & Case 4 & Case 5 & \multicolumn{1}{c}{Mean}
        & Case 1 & Case 2 & \multicolumn{1}{c}{Mean} \\
\midrule
\multicolumn{2}{l}{Total Length [m]}
& 136.391 & 295.630 & 232.400 & 196.798 & 345.779 & 241.400 
& 192.586 & 132.192 & 162.389\\
\midrule
Method & Configuration & \multicolumn{9}{c}{RMSE of ATE [m]}\\
\midrule
VINS‑MONO & Front(pinhole)   & 1.0525 & 1.4983 & 0.9452 & 2.1965 & 1.2426 & 1.3870 & 1.9443 & 1.5391 & 1.7417 \\
VINS‑MONO & Front            & 1.3605 & 1.1905 & 4.3665 & 2.5488 & 3.1643 & 2.5261 & 4.9033 & 3.7529 & 4.3281 \\
MCVIO     & Front,Rear       & 1.8731 & 2.4889 & 1.8822 & 3.6962 & 1.7403 & 2.3361 & 2.1654 & 1.6217 & 1.8936 \\
MCVIO     & Omnidirectional  & 1.7298 & 3.0179 & 1.7746 & 4.1562 & 2.3962 & 2.6149 & 2.9693 & 1.9459 & 2.4576 \\
MFVINS    & Front,Rear       & 0.7615 & 0.7316 & 0.8365 & 1.0311 & 1.5797 & 0.9881 & \underline{0.7630} & \underline{1.1729} & \underline{0.9680} \\
MFVINS    & Omnidirectional  & 0.6538 & 0.5627 & 0.6511 & 1.2697 & 0.5823 & 0.7439 & \textbf{0.6519} & \textbf{0.4588} & \textbf{0.5554} \\
MFVINS(D) & Front,Rear       & \underline{0.2243} & \underline{0.3419} & \underline{0.3552} & \underline{0.3579} & \underline{0.3523} & \underline{0.3263} & – & – & – \\
MFVINS(D) & Omnidirectional  & \textbf{0.2208} & \textbf{0.1915} & \textbf{0.2470} & \textbf{0.1767} & \textbf{0.2208} & \textbf{0.2114} & – & – & – \\
\bottomrule
\end{tabular}}
\caption{Localization results for different methods and configurations on the \textit{DRIVING} and \textit{OCCLUSION} sequences. Bold/underlined values indicate the best and second-best results, respectively.}
\label{tab:combined_result}
\end{table*}

\begin{figure*}
\centering
\includegraphics[width=1\textwidth]{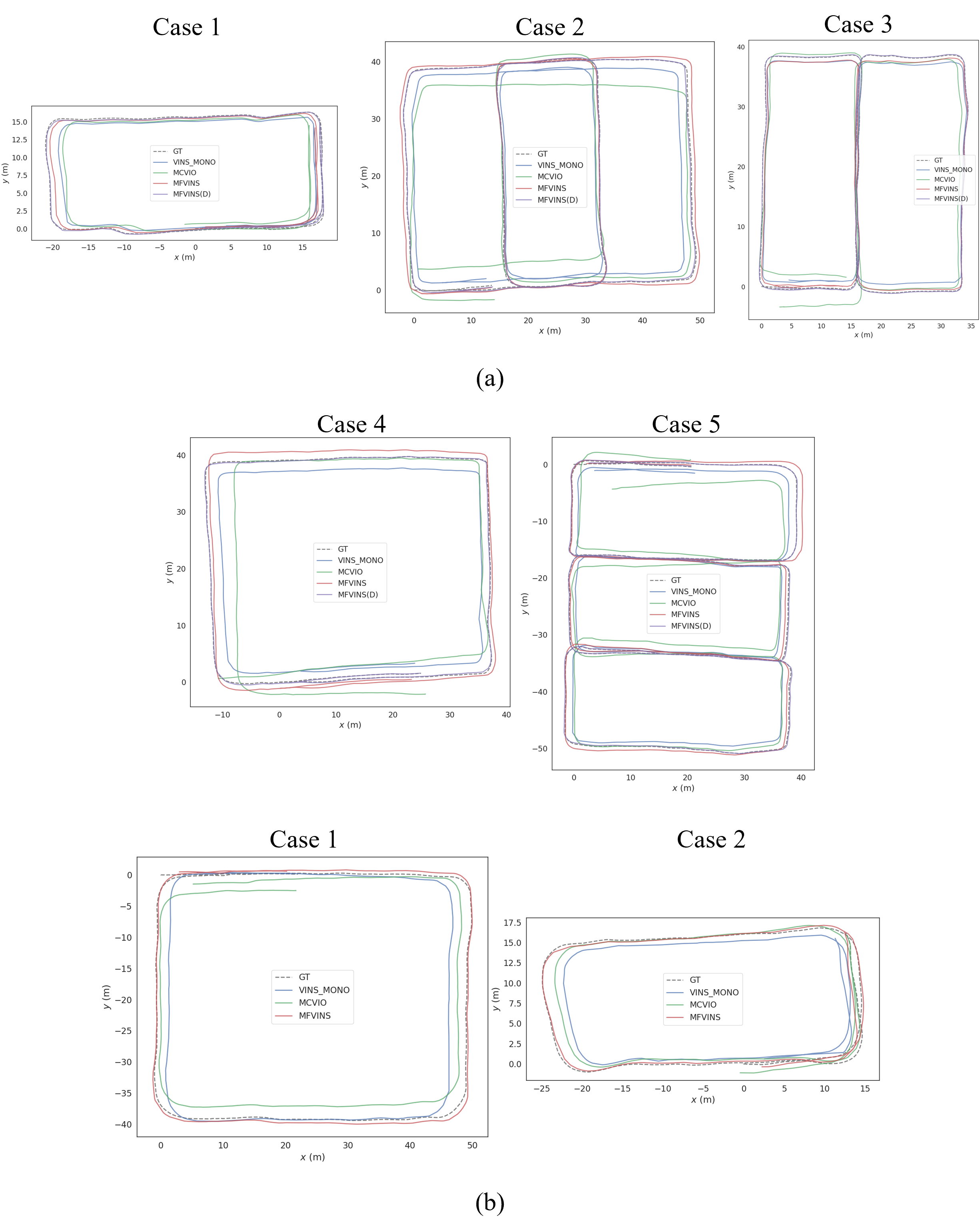}

\caption{
(a) and (b) show qualitative results of trajectory estimation on the \textit{DRIVING} and \textit{OCCLUSION} sequences, respectively. The dashed line denotes the ground-truth trajectory. The blue represents VINS-Mono, green represents MCVIO, red represents MFVINS, and purple represents MFVINS(D).
}
\label{fig:drving_result_qual}
\end{figure*}

\subsection{Experiments on \textit{DRIVING} Sequences}
To assess the performance of the proposed method in typical driving scenarios, we collected five different sequences and compared with other previous methods. For the monocular camera-based algorithm (VINS-MONO), we used different types of lenses to compare performance. For multiple camera-based algorithms (MCVIO and MFVINS), we tested setups using only front and rear cameras, as well as an omnidirectional camera configuration, to examine how the number of cameras affects performance. We also performed an experiment to investigate the effect of using a learning-based depth estimation method. When learning-based depth estimation was applied, we refer to the results as MFVINS(D). The quantitative results of these experiments are shown in Table 2. It can be seen that our proposed method achieves the most accurate localization. Using an omnidirectional camera configuration yields better results than using only front and rear cameras. Fig. \ref{fig:drving_result_qual}(a) presents qualitative results. For clarity, only results with a pinhole lens are shown for the monocular camera-based method (VINS-MONO), and only the omnidirectional camera configuration is shown for the multiple camera-based methods (MCVIO, MFVINS).

\subsection{Experiments on \textit{OCCLUSION} Sequences}

\begin{figure}
\centering
\includegraphics[width=0.8\textwidth]{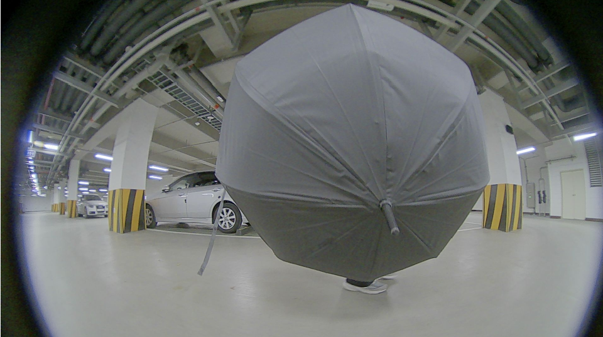}
\caption{A sample image of occlusion scenarios}
\label{fig:occ_ex}
\end{figure}

To evaluate performance under occlusion scenarios, we collected two sequences in which occlusion persisted continuously during driving, as illustrated in the top left image of Fig. \ref{fig:occ_ex}. The purpose of these experiments was to verify the robustness of the proposed system. Thus, depth estimation was not used in these experiments.

The quantitative results of these experiments are also shown in Table 2. Our proposed method demonstrates the highest accuracy for localization even under persistent occlusion. Additionally, the omnidirectional camera configuration is more robust than the front-and-rear-only setup in occlusion scenarios. The qualitative results are presented in Fig. \ref{fig:drving_result_qual}(b). 

\subsection{Ablation Study}
To verify the performance gains provided by each component of the proposed method, we conducted an ablation study. First, we examined how different feature extraction algorithms affect the processing time and the number of extracted features. Next, we evaluated whether using an IMU-aided feature tracking algorithm influences localization accuracy. Finally, we investigated the impact of applying a filtering algorithm to remove outlier features caused by fisheye distortion. Since the effects of a learning-based depth estimation method are already covered in Section 5.2, they are omitted here. All experiments were conducted using an omnidirectional camera configuration. For precise results of the contribution of each algorithm, the evaluation was performed on the trajectory before applying pose graph optimization.

\begin{table}[t]
\centering
\small
\resizebox{0.8\textwidth}{!}{%
\begin{tabular}{lcccc}
\toprule
METHOD &\makecell{Feature\\detection\\algorithm} & CPU/GPU &
\makecell{Running time\\per camera\\$[$ms$]$} &
\makecell{Number of\\untracked\\features} \\
\midrule
VINS‑Mono & Shi–Tomasi & CPU & 9.83 & \makecell{\underline{5,254}\\\underline{35.05\%}} \\
MCVIO     & VPI‑Harris & GPU & \textbf{0.35} & \makecell{97,322\\97.39\%} \\
MFVINS    & FAST       & CPU & \underline{0.37} & \textbf{\makecell{3,520\\32.58\%}} \\
\bottomrule
\end{tabular}
}
\caption{Performance of feature detection algorithm. MFVINS uses only CPU, but the running time per camera is similar to that using GPU, and the number of untracked feature is lower.}
\label{tab:feature_runtime}
\end{table}

\subsubsection{Feature Detection Algorithm}

The experiment on feature detection was conducted using ``Case 1" of the \textit{DRIVING} sequences. The evaluation results for the feature extraction time and the number of untracked features are presented in Table 3. The Shi-Tomasi algorithm used in the VINS-Mono system exhibited the longest processing time, indicating its unsuitability for multi-camera systems. In contrast, the VPI-Harris algorithm used in MCVIO offers the advantage of the shortest processing time by leveraging GPU acceleration to reduce CPU usage. However, it suffers from the drawback of excessive overlapping feature detections, which leads to a significant increase in the number of untracked features. This negatively impacts localization performance and, as shown in Table 2, explains the performance degradation observed in MCVIO when using omnidirectional cameras compared to using front and rear cameras only.

In the proposed method, the FAST feature extraction algorithm was employed to reduce computation time, demonstrating its suitability for multi-camera systems. It also resulted in the lowest number and ratio of untracked features, confirming its effectiveness in ensuring stable feature management and providing a favorable environment for omnidirectional camera-based localization.

\begin{figure}
\centering
\includegraphics[width=0.9\textwidth]{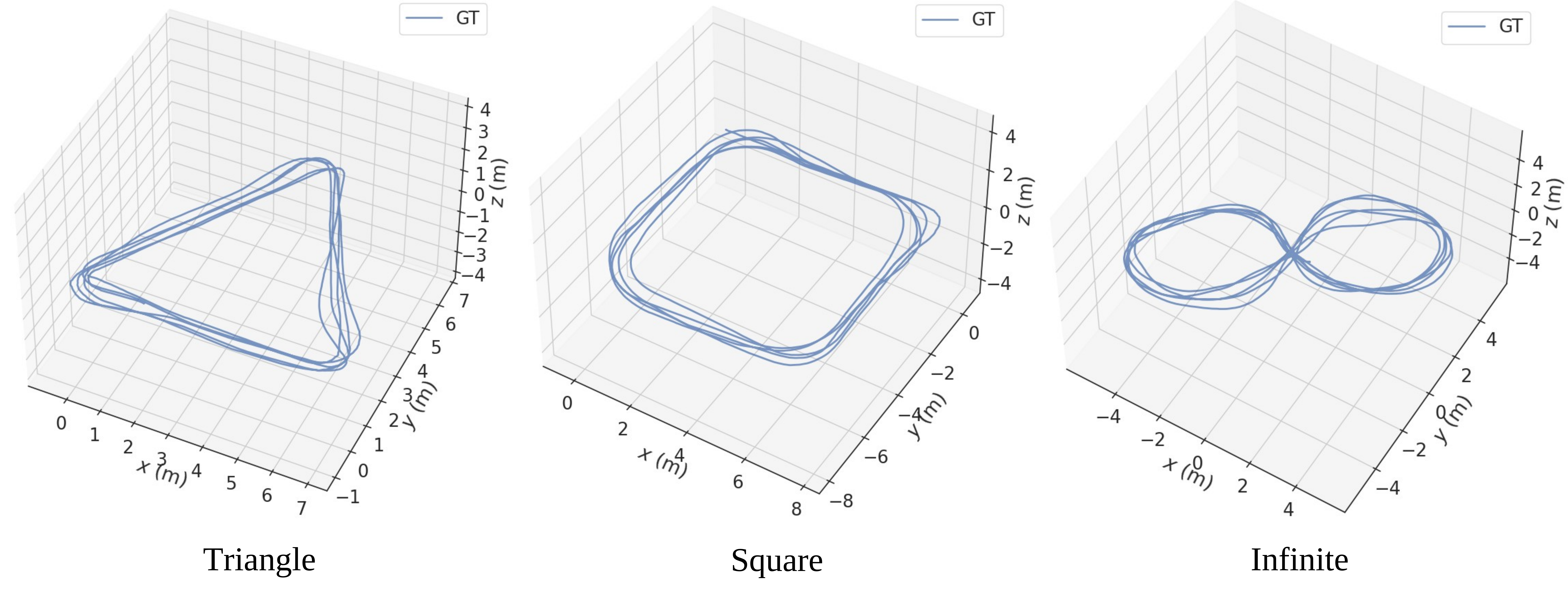}
\caption{Trajectories of \textit{ROTATIONAL} sequence following triangle, square, and infinite-shaped paths.}
\label{fig:rotational dataset}
\end{figure}

\begin{table}[t]
\centering
\small
\resizebox{0.6\textwidth}{!}{%
\begin{tabular}{lccc}
\toprule
\multicolumn{1}{l}{\textit{ROTATIONAL}} & Triangle & Square & Infinite \\
\cmidrule{1-4}
\multicolumn{1}{l}{Total Length [m]} & 119.51 & 139.77 & 157.56 \\
\midrule
IMU‑aided tracker &  \multicolumn{3}{c}{RMSE of ATE [m]}  \\
\midrule
 \multicolumn{1}{c}{X} & 0.5607 & 0.9846 & 0.5724 \\
 \multicolumn{1}{c}{O} & \textbf{0.4862} & \textbf{0.9228} & \textbf{0.5479} \\
\midrule
 IMU‑aided tracker &  \multicolumn{3}{c}{avg. number of tracked frames}  \\
\midrule
\multicolumn{1}{c}{X} & 16.3122 & 18.7986 & 17.3838 \\
\multicolumn{1}{c}{O} & \textbf{16.3383} &\textbf{18.8622} & \textbf{17.7092} \\
\bottomrule
\end{tabular}
}
\caption{Comparison results of IMU-aided feature tracking algorithm for \textit{ROTATIONAL} sequences.}
\label{tab:ablation_imu_aided}
\end{table}

\begin{table*}[t]
\centering
\small
\resizebox{1\textwidth}{!}{%
\begin{tabular}{lcccccccc}
\toprule
Sequence & \multicolumn{5}{c}{\textit{DRIVING}} & \multicolumn{2}{c}{\textit{OCCLUSION}} \\
\cmidrule(lr){2-6}\cmidrule(lr){7-8}
& Case 1 & Case 2 & Case 3 & Case 4 & Case 5 & Case 1 & Case 2 \\
\midrule
Total Length [m] & 136.39 & 295.63 & 232.40 & 196.79 & 345.77 & 192.58 & 132.19 \\
\midrule
Outlier Filtering & \multicolumn{7}{c}{RMSE of ATE [m]} \\
\midrule
\multicolumn{1}{c}{X} & 1.1153 & 2.4120 & 1.7342 & 2.5005 & 1.8269 & 1.7840 & 0.9781 \\
\multicolumn{1}{c}{O} & \textbf{0.9659} & \textbf{2.3840} & \textbf{1.6946} & \textbf{2.1537} & \textbf{1.4079} & \textbf{1.6919} & \textbf{0.9642} \\
\bottomrule
\end{tabular}
}
\caption{Comparison results of feature outlier filtering algorithm for all cases.}
\label{tab:outlier_filtering}
\end{table*}

\subsubsection{IMU-aided Feature Tracker}
Evaluating the performance of the IMU-aided feature tracking algorithm in the \textit{DRIVING} and \textit{OCCLUSION} sequences is challenging, since rotation effects are not pronounced in these scenarios. Therefore, to verify the effect of the algorithm, we additionally performed experiments with driving paths of triangular, square, and infinity-shaped that feature prominent rotational motion. The reference trajectories for these sequences are illustrated in Fig. \ref{fig:rotational dataset}.

The quantitative results of the experiment are presented in Table 4. When the proposed algorithm was applied, the localization accuracy was improved compared to the case without it for all three types. In addition, by leveraging inertial data from the IMU, the average number of frames per tracked feature increased, indicating that features could be tracked more stably over a longer period.

\subsubsection{Filtering Method for Distortion-included Feature Outlier}
Experiments were carried out on the \textit{DRIVING} and \textit{OCCLUSION} sequences, and Table 5 presents the quantitative results. The findings confirm that outliers caused by distortion adversely affect the accuracy of localization, underscoring the importance of filtering.

\section{Conclusion}
In this paper, we proposed a new visual-inertial system called MFVINS that utilizes multiple fisheye cameras. Building on VINS-Mono and MCVIO, we employed a grid-based FAST feature extraction assisted by an inertial measurement unit (IMU), as well as a distance-based filtering method to remove outlier features resulting from fisheye distortion. In addition, a learning-based depth estimation method was introduced to enhance depth accuracy, and pose graph optimization was used to reduce accumulated errors. Experiments were conducted on real-world datasets demonstrated that the proposed system outperforms existing monocular and multi-camera algorithms in terms of both accuracy and robustness. In future work, we plan to further enhance system practicality by optimizing sensor configurations and improving real-time processing, thereby maximizing the benefits of multi-camera setups and achieving even more accurate and robust perception capabilities.

\bibliographystyle{plainnat}

\end{document}